\documentclass[letterpaper]{article} % DO NOT CHANGE THIS
\usepackage{aaai25}  % DO NOT CHANGE THIS
\usepackage{times}  % DO NOT CHANGE THIS
\usepackage{helvet}  % DO NOT CHANGE THIS
\usepackage{courier}  % DO NOT CHANGE THIS
\usepackage[hyphens]{url}  % DO NOT CHANGE THIS
\usepackage{graphicx} % DO NOT CHANGE THIS
\usepackage{natbib}  % DO NOT CHANGE THIS AND DO NOT ADD ANY OPTIONS TO IT
\usepackage{caption} % DO NOT CHANGE THIS AND DO NOT ADD ANY OPTIONS TO IT
\usepackage{algorithm}
\usepackage{algorithmic}
\usepackage{amsmath,amsfonts}
\usepackage{mathtools}
\usepackage{multirow}
\usepackage{comment}
\usepackage[capitalize,noabbrev]{cleveref}

\usepackage{newfloat}
\usepackage{listings}
\DeclareCaptionStyle{ruled}{labelfont=normalfont,labelsep=colon,strut=off} % DO NOT CHANGE THIS
\floatstyle{ruled}
\newfloat{listing}{tb}{lst}{}
\floatname{listing}{Listing}
\title{FedAA: A Reinforcement Learning Perspective on Adaptive Aggregation for\\ Fair and Robust Federated Learning}
\author{
    Jialuo He\textsuperscript{\rm 1,\rm 2},
    Wei Chen\textsuperscript{\rm 3},
    Xiaojin Zhang\textsuperscript{\rm 1,\thanks{Corresponding author.}}
}
\affiliations{
    \textsuperscript{\rm 1} National Engineering Research Center for Big Data Technology and System\\ Services Computing Technology and System Lab, Cluster and Grid Computing Lab\\School of Computer Science and Technology, Huazhong University of Science and Technology, Wuhan, 430074, China\\ 
    \textsuperscript{\rm 2} School of Microelectronics and Communication Engineering, Chongqing University, Chongqing, 400044, China \\ 
    \textsuperscript{ \rm 3} School of Software Engineering, Huazhong University of Science and Technology, Wuhan, 430074, China\\
    xiaojinzhang@hust.edu.cn
 
}

\usepackage{bibentry}
\begin{document}

\maketitle

\begin{abstract}
Federated Learning (FL) has emerged as a promising approach for privacy-preserving model training across decentralized devices. However, it faces challenges such as statistical heterogeneity and susceptibility to adversarial attacks, which can impact model robustness and fairness. Personalized FL attempts to provide some relief by customizing models for individual clients. However, it falls short in addressing server-side aggregation vulnerabilities. We introduce a novel method called \textbf{FedAA}, which optimizes client contributions via \textbf{A}daptive \textbf{A}ggregation to enhance model robustness against malicious clients and ensure fairness across participants in non-identically distributed settings. To achieve this goal, we propose an approach involving a Deep Deterministic Policy Gradient-based algorithm for continuous control of aggregation weights, an innovative client selection method based on model parameter distances, and a reward mechanism guided by validation set performance. Empirically, extensive experiments demonstrate that, in terms of robustness, \textbf{FedAA}  outperforms the state-of-the-art methods, while maintaining comparable levels of fairness, offering a promising solution to build resilient and fair federated systems. Our code is available at https://github.com/Gp1g/FedAA.
\end{abstract}


\section{Introduction}
Federated learning (FL) is an emerging paradigm that enables collaborative model training while protecting the privacy of each participant's data \citep{fedavg, fedreview, reviewpfl, fedreview2}. Despite its potential, FL faces challenges stemming from data heterogeneity across clients and susceptibility to malicious attacks. These factors can lead to suboptimal global models that favor certain clients over others or models that are not robust to adversarial behaviors, undermining the core principles of FL. 

Personalized FL (PFL), represented by Ditto and lp-proj \citep{ditto, lp_proj}, emerged as a response to these challenges, tailoring models to individual client characteristics \citep{per1, reviewpfl}. Nonetheless, personalization typically focuses on the client level and does not adequately mitigate risks during the server-led aggregation phase. Similarly, existing works that incorporate notions of robustness \citep{robust1, robust2, robust3, robust4, fedmgda} and fairness \citep{qffl, fairfed, ERM} often do so separately, lacking an integrated approach that provides both concerns simultaneously.

In this paper, we present a novel method called Federated Adaptive Aggregation (FedAA), which employs deep reinforcement learning (DRL) to dynamically adjust the influence of each client's update during aggregation, thus balancing robustness and fairness at the server level. Our contributions are three-fold: 
\begin{itemize}
    \item Firstly, we propose a novel FL framework, FedAA, employing DRL to enhance both robustness and fairness via dynamically optimizing client contributions in FL. 
    \item Secondly, we conduct comprehensive experiments to validate the efficacy of the FedAA model. The results demonstrate significant improvements in robustness and maintain comparable levels of fairness against state-of-the-art (SOTA) methods. 
    \item Lastly, we perform ablation studies to pinpoint contributions of key components to model performance and provide insights into FedAA's fairness mechanisms, particularly in handling diverse client data distributions.
\end{itemize}

The remainder of this paper is organized as follows: First, we review the related works. Then, we provide a detailed description of the proposed FedAA framework, including the client selection algorithm, the DDPG-based optimization process, and the reward formulation. Next, we present our experimental setup and discuss the results. Finally, we conclude with suggestions for future research directions.

\section{Related Work}\label{Related Work}
\textbf{Robustness in Federated Learning.}
The robustness of FL models is a critical area of research, given the potential for adversarial interactions within decentralized training environments. Adversarial attacks, such as data poisoning and model update poisoning, have been identified as significant threats to the integrity of FL systems. Data poisoning introduces false information into the training datasets \citep{d_p1, d_p2, d_p3, d_p4, d_p5, d_p7, d_p6}, while model update poisoning, one is Byzantine attacks, involves an $\alpha$-fraction (typical $\alpha<0.5$) of clients acting maliciously to disrupt the learning process. Various Byzantine robust SGD methods have been proposed to mitigate these threats to enhance the resilience of FL models against such attacks \citep{robust1, robust2, robust3}. This paper aligns with the robustness definition by \citet{ditto} and considers the following attack models for evaluation:

\textbf{Definition 1} (Robustness). 
In the case of a certain Byzantine attack, if model $w_1$ achieves higher mean test accuracy across benign clients compared to model $w_2$, then we say that model $w_1$ is more robust than model $w_2$. We employ three common attack methods to evaluate the robustness of our model. We use $\mathbf{\Tilde{w}}_k$ to represent the malicious messages sent by client $k$.

\noindent $\bullet$ \textbf{Same-value attacks}: Malicious client $k$ sends parameters can be denoted as $\mathbf{\Tilde{w}}_k=m\mathbf{1}$, where $m \sim \mathcal{N}(0, \tau^2)$ represents the intensity of attack, $\mathbf{1}$ is a vector of ones, with the same size of parameters as the benign clients.

\noindent $\bullet$ \textbf{Sign-flipping attacks}: Malicious client $k$ sends sign-flipped and scaled messages, which can be represented as $\mathbf{\Tilde{w}}_k=-|m|\mathbf{\Tilde{w}}'_k$, where $\mathbf{\Tilde{w}}'_k$ denotes the correct updates, $m \sim \mathcal{N}(0, \tau^2)$ represents the intensity of attack.

\noindent $\bullet$ \textbf{Gaussian attacks}: The messages sent by Byzantine client $k$ follow a Gaussian distribution, which can be formulated as $\mathbf{\Tilde{w}}_k \sim \mathcal{N}(\mathbf{0},\tau^2\mathbf{I})$.

\textbf{Fairness in Federated Learning}
Fairness in the context of FL is a multifaceted issue that encompasses performance fairness, collaboration fairness, and model fairness \citep{fairfed1}. This paper focuses on performance fairness, which aims to ensure that the model performs well across diverse client datasets, thereby preventing any client with less common data from being disadvantaged. The concept of fairness is influenced by the work of \citet{ditto}, which advocates for a fair model to provide an equitable distribution of performance across all clients: 

\textbf{Definition 2} (Performance Fairness). In the case of a heterogeneous federated network, if model $w_1$ achieves a lower standard deviation (std) of test performance across $N$ clients than model $w_2$, i.e., \textrm{std} $\{F_k(w_1)\}_{k\in [N]}$ $ <$  \textrm{std} $ \{F_k(w_2)\}_{k\in [N]}$, where $F_k(\cdot)$ represents the test loss of client $k \in [N]$, then we say that model $w_1$ is more fair than model $w_2$.

Prior research has proposed reweighting techniques to address fairness, adjusting the contribution of client updates based on their performance \citep{qffl, fairfed, ERM}. However, these methods may come at the cost of mean test accuracy and may not be robust against adversarial attacks \citep{robust1, robust2, robust3, robust4, fedmgda}.

To provide robustness and fairness, several frameworks aim to tune the deviations between local and global models finely. The Ditto framework by \citet{ditto} address this by allowing controlled deviations from the global model to foster personalization, which contributes to fair and robust outcomes across diverse client datasets. Similarly, \citet{lp_proj} propose a projection method that manages these deviations by embedding local models within a shared low-dimensional subspace, thus enhancing communication efficiency while ensuring robustness against adversarial attacks and fairness in resource allocation. Building on these foundations, our approach integrates deep reinforcement learning (DRL) to dynamically optimize the aggregation process, focusing on achieving a robust and fair global model that adapts to real-time network changes, thus further personalizing client contributions effectively.

\textbf{Deep Reinforcement Learning.}
The integration of DRL into FL represents a promising frontier in addressing the challenges faced by FL. Notable examples include FAVOR \citep{favor}, which utilizes a deep Q-learning (DQN) algorithm for client selection \citep{e_r_t_n}, and FedRL \citep{fedrl}, which operates within a discrete action space. Yet, these approaches are often limited by their reliance on discrete action spaces and greedy policies, which may not be suitable for the complex, continuous action spaces inherent in FL. This paper explores the application of DRL, specifically the DDPG algorithm \citep{DDPG}, to handle continuous control in FL, offering a more nuanced approach to client aggregation.

\section{Methodology}\label{Methodology}
In this section, we first outline the foundational concepts of FL and DRL. Then we formulate the optimized function and provide the details of the proposed algorithm. 
\subsection{Preliminary}
DRL embodies a learning paradigm in which an agent learns to interact with its environment through a process of trial and error. In the context of DRL, at each timestep $t$, the agent perceives the current state $s(t)$ of the environment, selects an action $a(t)$, and subsequently receives a reward $r(t)$. This interaction leads to a transition to the next state $s(t + 1)$. The overarching goal of DRL is to identify a policy that maximizes the cumulative discounted reward, defined as $R=\sum_{t=1}^T \gamma^{t-1}r(t)$, where $\gamma$, the discount factor, is a value within the range (0, 1].

Our work introduces a DRL framework based on the Deep Deterministic Policy Gradient (DDPG) algorithm \citep{DDPG}, which extends the capabilities of traditional Q-learning methods to handle continuous action spaces. The DDPG algorithm operates using an actor-critic structure \citep{a-c}, comprising two neural networks: an actor network $\pi(s|\theta^\pi)$ that selects actions, and a critic network $Q(s, a|\theta ^Q)$ that evaluates the chosen actions. These networks are parameterized by $\theta^\pi$ and $\theta^Q$, respectively. To improve stability and performance, we implement experience replay, target actor network $\pi'(s|\theta^{\pi'})$, and the target critic network $Q'(s, a|\theta ^{Q'})$, which facilitate more consistent learning updates. 

The actor network aims to develop a policy that maximizes the expected return $J=\mathbb{E}_{r_i,s_i,a_i}[R]$ from the environment's start distribution. This policy refinement is achieved through gradient ascent, utilizing the gradient 
\begin{equation}\label{loss_actor}
    \nabla_{\theta^\pi} J \approx \frac{1}{N_d} \sum_i \nabla_{a}Q\big (s(i),\pi(s(i))|\theta^Q \big) \nabla_{\theta^\pi} \pi(s(i)|\theta^\pi),
\end{equation}
where $N_d$ is the batch size. The critic network's role is to approximate the action-value function $Q(s, a|\theta ^Q)$, which predicts the expected return for a given state-action pair. The critic's learning process involves minimizing the loss function:
\begin{equation}\label{loss_critic}
    L=\frac{1}{N_d} \sum_i \big(y(i) - Q(s(i), a(i)|\theta^Q)\big)^2,
\end{equation}
with $y(i)$ defined as the target value, 
\begin{equation}
     y(i)=r(i) + \gamma Q'\big(s(i+1),\pi'(s(i+1)|\theta^{\pi'})|\theta^{Q'}\big).
\end{equation}

The target networks are periodically updated using the soft update equation $\theta'\gets \varepsilon\theta+(1-\varepsilon)\theta'$, where $\varepsilon$ is a small positive constant \citep{DDPG}, ensuring that the target networks slowly track the primary networks and provide a stable target for the learning process.

\subsection{FedAA Overview}
We present Federated Adaptive Aggregation (FedAA), a framework designed to enhance server-level robustness and fairness in federated environments. Our approach streamlines the aggregation process by integrating a novel client selection algorithm that identifies the top M\% of clients based on model parameter proximity, thus protecting the aggregation against adversarial influences. The server, acting as a DDPG-driven agent, leverages this selection to determine the aggregation weights for these clients, guiding the global model update. FedAA's flexibility accommodates both full and partial client participation scenarios, enhancing its applicability in diverse federated learning contexts (\cref{ipmattack}).

Our methodology is designed to tackle the dual goals of optimizing global models while ensuring robustness and fairness on the server-side. We articulate this through a bi-level optimization that minimizes the aggregated local client objective while maximizing the aggregated model's accuracy on a fair validation set (see \textbf{Reward}). This is formalized as:
\begin{equation}
     \max \limits_{w_g}\enspace F_g(w_g)\coloneqq \text{Acc}(w_g,\mathcal{D}_g), 
\end{equation} 
where $w_g={\operatorname{argmin}_w\enspace} G(F_1(w),...F_N(w))$, $F_g$ denotes the global optimization problem, Acc$(w_g,\mathcal{D}_g)$ means the global model $w_g$ test accuracy on the dataset $\mathcal{D}_g$, $G(\cdot)$ represents the aggregation function at the server side, $N$ denotes the number of clients, $F_k$ is the local optimization problem for client $k$, i.e., $F_k \coloneqq \mathbb{E}_{x_k}[f(w,x_k)]$, $x_k$ is a random sample draw according to the distribution of client $k$, $f(w,x_k)$ is the local loss function. In this context, $w_g=\sum_{k=1}^{N} {a_k}\cdot w_k$, $a_k\ge0$ represents the aggregate weight for client $k$, $\sum_{k=1}^{N}a_k = 1$, and $w_k$ is the optimized local model of client $k$.

The reward, defined by the model's accuracy on a fair dataset $\mathcal{D}_g$, incentivizes the agent to pursue actions that lead to a balanced and robust global model.

\textbf{State.}\label{3.1.1} 
In the application of DRL to FL, the most straightforward idea is to consider the parameters of each client's model as states and input them into the actor network. However, when the client's model is a neural network, the parameter size becomes exceedingly large, rendering this idea unfeasible. 

\begin{algorithm}[!h]
   \caption{Client Selection}
   \label{alg_cs}
    {\bfseries Input:} client models $w_{all}=[w_1,...,w_N]$, an $N \times N$ distance matrix $\bf{C}$ filled with zeros, M\%.
   \\
    {\bfseries Output:} state $s$, top-$\rm M \%$ clients' model $w_{top}$.

   \begin{algorithmic}[1]
   \STATE Flatten parameters of each client model to $w'_1,...,w'_N$
   \FOR{$i=1$ {\bfseries to} $N$}
   \FOR{$j=1$ {\bfseries to} $N$}
   \STATE $\mathbf{C}_{i,j}= \| w'_i - w'_j  \|_2 $
   \ENDFOR
   \ENDFOR
   \STATE Summing the rows of matrix $\bf{C}$ and selecting the top-$\rm M \%$ rows with the minimum sums.
   \STATE The sum of selected $\rm M \%$ rows forms a distance vector $s=[d_m,\dots, d_{M}]$ as the state. 
\end{algorithmic}
\end{algorithm}

Inspired by the idea of FABA \citep{faba}, there is a strong likelihood that the Euclidean distance among the model parameters of benign clients is closer than the distance between the model parameters of benign clients and malicious clients. Based on this observation, we propose a novel method for client selection and apply it to the generation of states. The details are shown in \cref{alg_cs}. Through this algorithm, we can simultaneously obtain state $s=[d_m,\dots, d_{M}]$ and top-$\rm M \%$ clients' model $w_{top}=[w_m,...,w_{M}]$, $d_m$ is the sum of the distances between the client model $m$ and the remaining client model parameters. These clients are the $\rm M \%$ clients whose model parameters have the minimum total distance from the model parameters of all other clients. Notably, we perform normalization on the state $s$ to get a stable training process. Further, the reason we do not directly use the distance matrix $\mathbf{C}$ as input is that, if we consider 100 clients, the state will be in a 10,000-dimensional space, which would significantly increase computational costs. 

\textbf{Action.}
Compared to the FAVOR \citep{favor}, which constrains its action space to discrete numbers ranging from $1$ to $\rm{N}$, representing the IDs of selected clients, our proposed DDPG-based algorithm enjoys a continuous control feature. The actor network will generate a continuous action $a=[a_m,...,a_M], \sum_i a_i=1$ based on the input state $s$. In our algorithm, we set the action as the aggregation weights of selected top-$\rm M \%$ clients. 

\textbf{Reward.}
When the server aggregates the parameters of selected $\rm M \%$ clients, the server will get a reward $r$ as feedback. The actor network performs gradient ascent to optimize its action for a higher expected cumulative reward. Hence, the design of the reward will guide the optimization direction of the actor network. {Building a small dataset on the server side has many applications in both academia \citep{serverdata4, reviewpfl, serverdata, serverdata2, serverdata3, d_p7} and industry \citep{gboard}. For example, Google can allow its employees to use Gboard to obtain server-side server datasets for next-word prediction \citep{gboard}. For image recognition tasks like identifying cats and dogs, a group of people can be hired to label the cat and dog images. In the case of this paper, only a small dataset (e.g., 100 to 1000 training samples) is needed, and the service provider can usually afford the cost of manual collection and labeling. This dataset then can be used for the evaluation of the global model on various performance metrics.}
Here, we construct a fair held-out validation set at the server and use the test accuracy of the aggregated global model $w_g$ on this validation set as the reward. 

We adopt this approach for several reasons: testing at the server does not incur additional communication overhead, and it allows training an unbiased global model. Specifically, taking the example of MNIST, we construct a validation set at the server with 100 images for each digit, totaling 1000 images. Achieving higher rewards on such a validation set incentivizes the agent to make actions that are more fair to each client, unlike FedAvg, which assigns higher weights to clients with more images, which may lead to significant unfairness in a non-identically distributed (non-IID) setting. {Conversely, if our constructed dataset disproportionately features a high quantity of digits 0 and 1, while scantily representing other digits, in pursuit of obtaining higher rewards, the agent might be incited to discern which clients' datasets are richer in these particular digits, subsequently elevating their aggregation weights. Such a scenario inadvertently precipitates a disparity detrimental to the remaining clients (\cref{unfairdataset}).}

\subsection{Algorithm}
In this section, we provide a comprehensive overview of the DDPG-based training process. The optimization of FedAA is composed of two components: (i) within the DRL workflow, the agent updates its actor and target networks; (ii) clients solve their local problems. The details are shown in \cref{alg_FedAA}. 

\begin{algorithm}[!h]
   \caption{FedAA: Fair and Robust Federated Learning with Adaptive Aggregation}
   \label{alg_FedAA}
    {\bfseries Input:} $w_g(0)$, $\pi \left( s|\theta ^\pi \right)$, $Q\left( s,a|\theta ^Q \right)$, $\theta ^{\pi '}$, $\theta ^{Q'}$, ${\cal U}$, $T$, $N$, $R$.
\begin{algorithmic}[1]
   \STATE Server(Agent) sends global model $w_g(0)$ to all clients.
   \STATE $s(0), w_{top}(0)\gets$ ClientSelection$(w_{all},\bf{C}, \rm M)$ 
   \FOR{$t=0$ {\bfseries to} $T-1$}
   \STATE Server observes the state $s(t)$, and makes action $a(t)=\pi \left( {s(t)|{\theta ^\pi }} \right) +\cal N$ ($\cal N$ is an exploration noise).
   \STATE Update global model $w_g(t)\gets \sum_i a_i(t) \cdot w_{top_i}(t)$.
   \STATE $r(t)\gets$Evaluation$(w_g(t))$.
   \STATE Server sends $w_g(t)$ to all $N$ clients.
   \FOR{$k=1$ {\bfseries to} $N$}
   \STATE Client $k$ solves its local problem for $R$ rounds.
   \ENDFOR
   \STATE $s(t+1), w_{top}(t+1)\gets$ ClientSelection $(w_{all},\bf{C}, \rm M)$
   \STATE Store $(s(t),a(t),r(t),s(t+1))$ in the experience replay buffer $\cal U$.
   \STATE Sample a batch of experience from $\cal U$ to update $\theta^\pi,\theta^Q$, using \cref{loss_actor} and \cref{loss_critic}.
   \STATE Soft update $\theta^{\pi'},\theta^{Q'}$ via $\theta'\gets \varepsilon\theta+(1-\varepsilon)\theta'$.
   \ENDFOR
\end{algorithmic}
\end{algorithm}

In this context, we adopt DRL to acquire an aggregation function to trade off robustness and fairness. After initialization, the FedAA algorithm progresses through sequential steps. At each step $t$, the server (acting as an agent) observes the current state $s(t)$, derived through the process of $\textbf{Client Selection}$. It then makes a deterministic action and performs aggregation, thereby generating a new global model $w_g(t)$. Then, an evaluation of the fair held-out dataset serves as the reward $r(t)$. The server then broadcasts the updated global model to all clients, who then solve local subproblems for $R$ rounds. Following this training phase, a new round of $\textbf{Client Selection}$ takes place to obtain the next state $s(t+1)$ for the subsequent iteration. The acquired transition $(s(t),a(t),r(t),s(t+1))$ will be preserved in the replay buffer for subsequent network updates. The actor and critic networks update at every step $t$, while the target actor and critic networks update once every two steps (soft update). The slow-updating target networks provide a stable learning process \citep{DDPG}.

\textbf{Trade-off Between Robustness and Fairness.} In previous works \citep{ditto, lp_proj}, the local model is susceptible to collapse under strong attacks (such as sign flipping). This can be explained by the insufficient robustness of the global model. In contrast, our proposed FedAA can simultaneously offer robustness and fairness at the server level. Specifically, when there is a certain fraction $\alpha$ ($\alpha < 0.5$) of malicious clients present, we can control the percentage $\rm M\%$ of clients participating in aggregation to trade off robustness and fairness. The intuition is that: when we adopt a larger value of $\rm M$, it also increases the risk of introducing malicious clients during aggregation. However, simultaneously, introducing more client parameters in non-IID situations can enhance the generalization capability of the global model, implicitly promoting fairness across the clients.

\textbf{Selections of State, Action, and Reward.} Note that, within DRL, the design of $\textbf{states}$, $\textbf{actions}$, and $\textbf{rewards}$ is highly personalized and not standardized. It can differ depending on the specific objectives of different algorithms, allowing for various setups of states, actions, and rewards. The algorithm proposed in this paper, based on DDPG, only presents a framework capable of continuous control within the context of FL. There is considerable potential for further exploration and development in subsequent work.

\section{Numerical Experiments}\label{experiments}
In this section, we provide representative evaluation results to demonstrate that FedAA can achieve superior test accuracy, robustness, and comparable fairness compared to SOTA methods. We summarize the datasets, models, and other configurations used in this paper in Appendix A.1, and full results in Appendix A.2. We compare FedAA with two SOTA approaches in robustness and fairness, namely Ditto \citep{ditto}, lp-proj \citep{lp_proj}, and a baseline FedAvg \citep{fedavg} which are summarised in Appendix A.

We then exhibit the tradeoff capability between the robustness and fairness of FedAA. Next, we conduct supplementary experiments regarding the reward design, actual execution time, and partial participation to illustrate the feasibility of FedAA.

\begin{figure}[!h]
\centering
\includegraphics[width=\columnwidth]{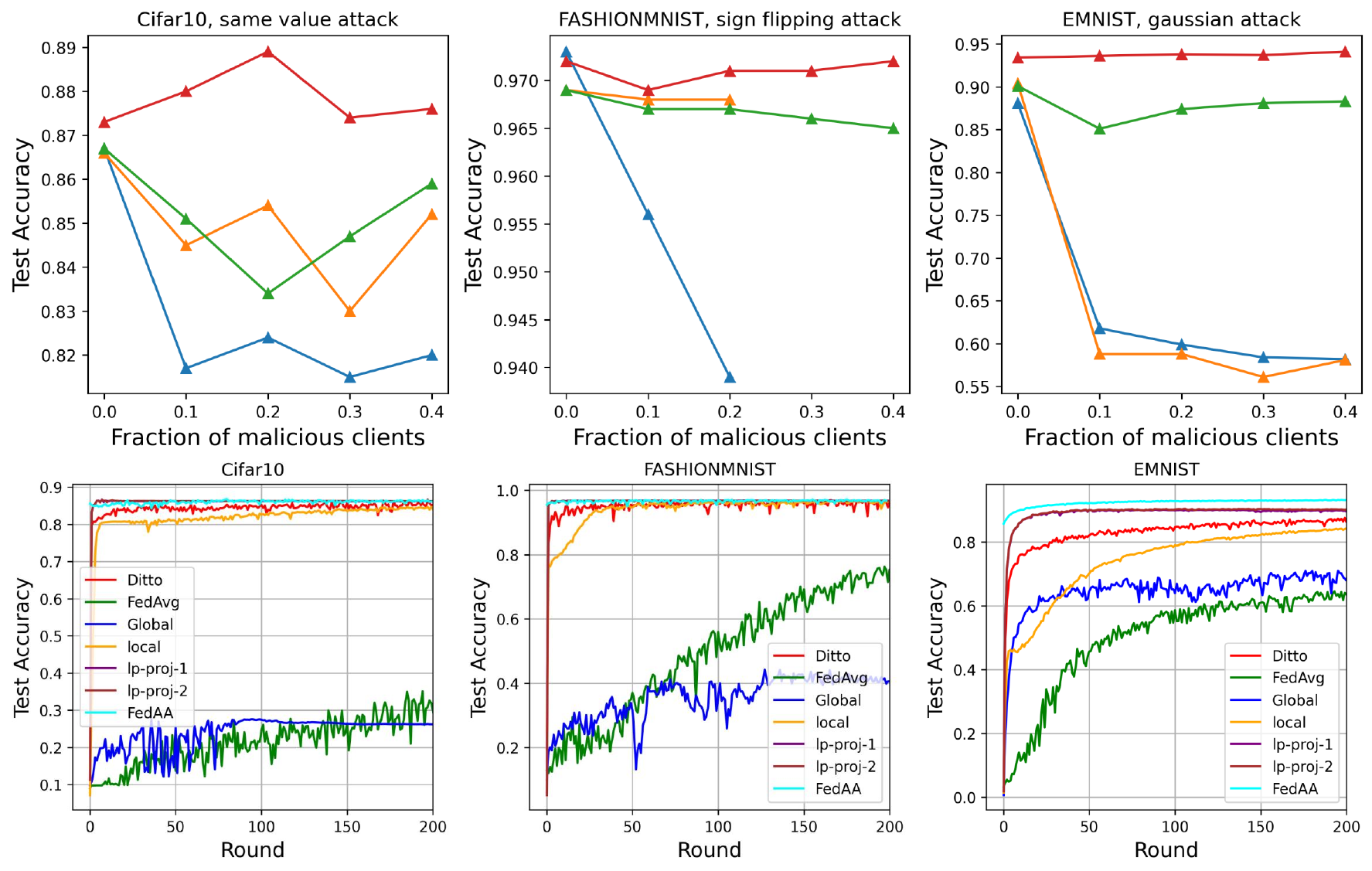}
\caption{The figures in the first line represent robustness performance (i.e. mean test accuracy across benign clients) of three different datasets subjected to three different attacks. The figures in the second line depict the performance of three different datasets with no malicious clients.}
\label{robustness}
\end{figure}

\textbf{Robustness and fairness.} Following the definition of robustness \citep{ditto}, we provide empirical results on three different datasets, under three different attacks: same-value attack, sign-flipping attack, and Gaussian attack, with the parameter $\tau$ set to $\{100, 10, 100\}$ respectively. In the absence of malicious clients, \cref{robustness} demonstrates that FedAA achieves performance similar to SOTA methods on CIFAR10 and EMNIST datasets. However, it shows a slight inferiority compared to Ditto \citep{ditto} on FASHIONMNIST with no adversaries. Furthermore, under the same value attack, the performance of Ditto and lp-proj \citep{lp_proj} slightly decreases with the increasing number of malicious clients. While FedAA shows a slight improvement. Additionally, under two other types of attacks, Ditto and lp-proj-2 perform poorly. Specifically, subjected to Gaussian attacks, both algorithms exhibit a notable deterioration in performance. In the case of a strong attack, i.e. sign flipping, Ditto and lp-proj-2 collapse when the fraction of malicious clients exceeds 0.2.

\begin{figure}
  \includegraphics[width=\columnwidth]{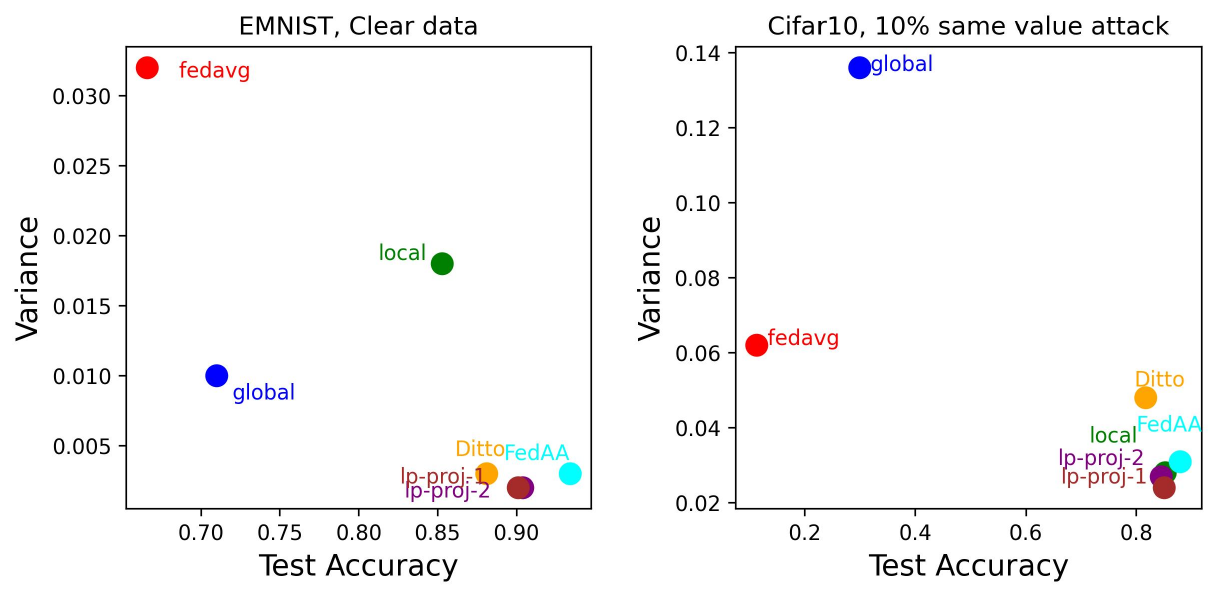} 
  \caption{The tradeoff between test accuracy and fairness within different methods. The closer the approach is to the lower right corner, the better.} 
  \label{acc and fairness}  
\end{figure}

The tradeoff between test accuracy and variance for different baselines is illustrated in \cref{acc and fairness}. We have examined two scenarios: one involving a malicious client and one without. As shown in \cref{acc and fairness}, FedAA achieves superior accuracy. However, its variance is marginally higher compared to other approaches. This can be mitigated by tuning $\rm M$ which will be discussed later.  

Numerous experimental results demonstrate that FedAA can provide robustness and fairness, and due to the space limitation, we show full results in Appendix A.2. Further, we present some analysis of the underlying reasons. As mentioned in Ditto \citep{ditto}, superior results can be achieved through the execution of local fine-tuning for 50 epochs on the global model after specific communication rounds. However, determining the optimal 'point' for early stopping during training poses a challenge, especially in the presence of a fraction of malicious clients corrupting the global model. In contrast, FedAA can constantly provide a relatively robust global model through a robust client selection algorithm. Meanwhile, the server, also referred to as the agent, aims to maximize the expected accumulative reward. It optimizes the aggregation weights in each round and learns a policy to make better decisions. In non-IID scenarios, an aggregation function that involves interaction with clients and incorporates feedback for continuous learning demonstrates superior performance compared to FedAvg \citep{fedavg}, which determines aggregation weights simply based on the sample size in each round.

\begin{figure*}[!h]
\begin{center}
\centerline{\includegraphics[width=\textwidth]{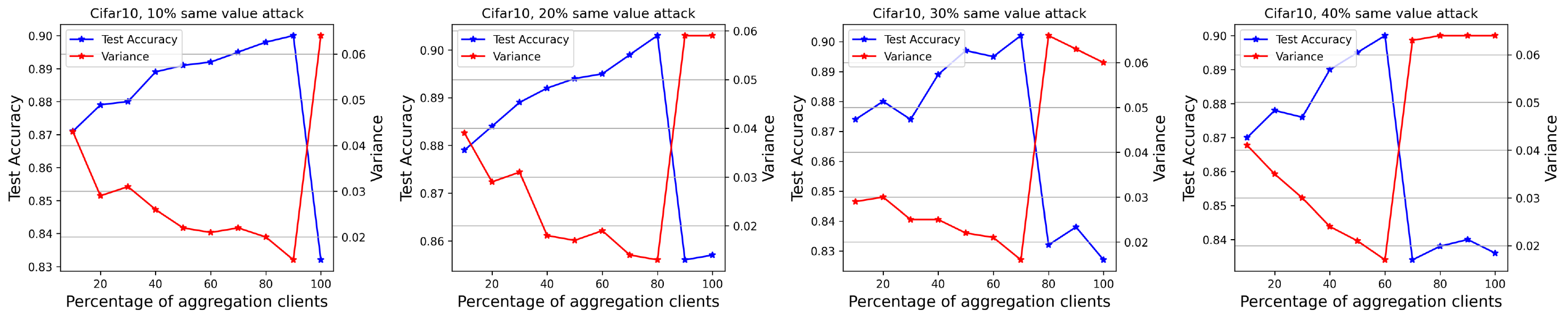}}
\caption{The performance and tradeoff between robustness and fairness of different $\rm M$ (The numbers on the x-axis in the figures represent the corresponding $\rm M\%$, e.g. 80 means $\rm M=80\%$).}
\label{tradeoff}
\end{center}
\end{figure*}

\textbf{Tradeoff between robustness and fairness in FedAA.} Results are shown in \cref{tradeoff}. Experiments are conducted on CIFAR10, while in the presence of different percentages of malicious clients, all subjected to the same value attack. We set the number of clients participating in aggregation, $\rm M$, ranging from 10\% to 100\%. Specifically, there exists a certain threshold, i.e. if there are 100 clients with 20\% being malicious clients, then the threshold is set at 80. We see that, under different percentages of malicious clients, there is a certain pattern in the changes of test accuracy and variance. Specifically, as the value of $\rm M $ continuously increases, test accuracy initially oscillates upward, reaching its maximum at the threshold, and then experiences a sharp decline. In contrast, variance exhibits the opposite pattern, with a continuous increase in $\rm M $ leading to initial oscillations downward, reaching its minimum at the threshold, and then rapidly rising. Taking the example of 20\% malicious client, under the same value attack, $\rm M=80\%$ achieves the highest mean test accuracy of 90.3\% and the lowest variance of 0.013 (complete results are available in Appendix A.2). The reason for not optimizing $\rm M$ stems from the associated risks. Consider a scenario in which malicious clients collaborate (see \citep{ipm}). These clients might submit regular updates for $t$ epochs, during which time M could be optimized to an excessively high value, potentially reaching 100\%. Should these malicious clients then collectively submit anomalous updates, they could easily corrupt the entire FL system.

The experimental results validate the initial proposition of FedAA, indicating that we can provide a tradeoff between robustness and fairness by controlling the number of participating clients $\rm M$ in aggregation. To elaborate further, when the server encounters an attack at first, it may receive lower rewards. In subsequent network updates, the server learns to assign lower aggregation weights to clients suspected of being malicious, thereby enhancing robustness. Additionally, in order to attain higher rewards, the server also learns how to allocate weights among benign clients to ensure fairness.

\begin{figure}
    \includegraphics[width=\columnwidth]{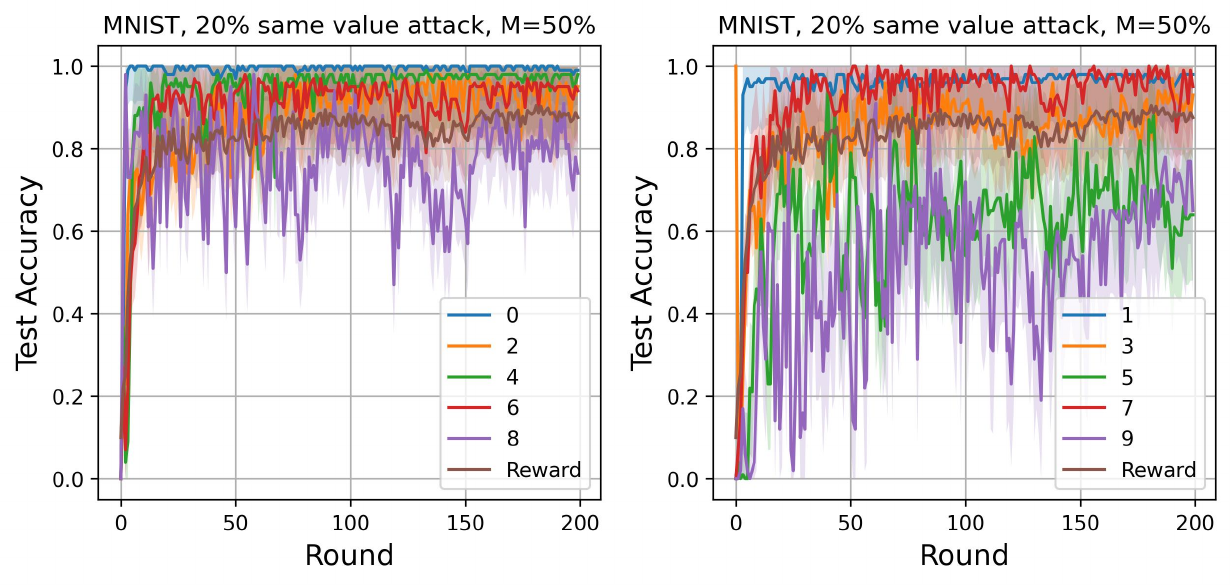}
    \caption{The convergence curve of reward $r$ of each class at the server.} % under the same value attack.} 
\label{reward}
\end{figure}
\textbf{Reward design.} \cref{reward} illustrates the effectiveness of carefully designed rewards. The figure depicts the convergence process in test accuracy for each digit. To enhance clarity, we separate odd and even digits into two separate figures. From \cref{reward}, it is evident that most digits converge to similar ranges, except for digits 5 and 9, which converge around 65\%. These results are deemed fair for the majority of clients. However, it presents a relative unfairness for clients primarily composed of digits 5 and 9. This issue can be mitigated by refining the design of rewards in future work.

\textbf{Compression methods and excution time.} We compared two compression methods along with their corresponding actual execution times. One is reducing the number of neurons in the hidden layers of the DDPG network to 256, 128, and 64, respectively. The other method entailed selecting only the parameters of the last hidden layer (LHL) of the client model, instead of all layers (AL). 
\begin{table}[htbp]
\caption{Compression methods and actual execution time.}
\label{compress}
\begin{center}
\begin{sc}
\resizebox{\columnwidth}{!}{
\begin{tabular}{cccc} 
\hline
dataset                       & methods     & Runtime & Acc           \\ 
\hline
\multirow{3}{*}{Mnist}        & FedAA(AL 256) & 5,173s  & 0.978(0.001)  \\
                              & FedAA(AL 128) & 5,127s  & \textbf{0.979(0.001)}  \\
                              & FedAA(AL 64)  & 5,103s  & 0.978(0.000)  \\ 
\hline
\multirow{2}{*}{FASHIONMNIST} & FedAA(AL 256) & 19,872s & \textbf{0.975(0.038)}  \\ 
                              & FedAA(LHL)    & 18,314s & 0.974(0.038)  \\
\hline
\multirow{2}{*}{CIFAR10}      & FedAA(AL 256) & 14,392s & \textbf{0.875(0.024)}  \\
                              & Ditto       & 13,096s & 0.820(0.042)  \\
\hline
\end{tabular}}
\end{sc}
\end{center}
\end{table}

In \cref{compress}, AL 256 indicates uploading all parameters of the client model, with the hidden layer neuron count in the DDPG network set to 256. As can be seen in \cref{compress}, the impact of different hidden layer dimensions in the actor and critic networks within DRL on test accuracy is limited. We delve further into applying another compression technique and measure the real-world execution time of FedAA, as is elaborated in \cref{compress}. For clarification, the term 'last hidden layer' specifically refers to utilizing the parameters of the model's final hidden layer as input for the algorithm described in \cref{alg_cs} \citep{dim_reduce}. The employed compression strategy is effective, yielding robust and competitive results. Although FedAA exhibits a slightly increased operational time in comparison to Ditto, the enhanced performance by 5.5\% justifies the additional duration, which is deemed manageable.

\begin{table*}[htbp] 
\caption{ Comparison of FedAA and baseline methods under inner product manipulation (IPM) \citep{ipm} attack, where C indicates client participation ratio, across CIFAR-10 and MNIST datasets.}
\label{ipmattack}
\begin{center}
\begin{small}
\begin{sc}
\resizebox{0.6\textwidth}{!}{
\begin{tabular}{ccccc} 
\hline
                         &                     &                       & \multicolumn{2}{c}{IPM}                        \\ 
\cline{4-5}
Dataset                  & Methods             & clear                 & 10\%                  & 20\%                   \\ 
\hline
\multirow{5}{*}{cifar10} & FedAvg              & 0.377(0.048)          & 0.105(0.045)          & 0.114(0.039)           \\
                         & Ditto               & 0.867(0.019)          & 0.857(0.022)          & 0.849(0.023)           \\
                         & lp-proj-1           & 0.867(0.021)          & 0.864(0.029)          & 0.857(0.031)           \\
                         & FedAA(C=100\% M=30\%) & 0.873(0.015)          & \textbf{0.870(0.017)} & 0.861(0.018)           \\
                         & FedAA(C=50\% M=30\%)  & \textbf{0.878(0.013)} & 0.869(0.016)          & \textbf{0.862(0.018)}  \\ 
\hline
\multirow{5}{*}{mnist}   & FedAvg              & 0.904(0.014)          & 0.425(0.006)          & 0.415(0.006)           \\
                         & Ditto               & 0.972(0.005)          & 0.937(0.011)          & 0.933(0.012)           \\
                         & lp-proj-1           & 0.971(0.007)          & 0.969(0.008)          & 0.968(0.008)           \\
                         & FedAA(C=100\% M=30\%) & 0.977(0.002)          & \textbf{0.977(0.002)} & 0.975(0.004)           \\
                         & FedAA(C=50\% M=30\%)  & \textbf{0.984(0.001)} & 0.976(0.002)          & \textbf{0.977(0.003)}  \\
\hline

\end{tabular}}
\end{sc}
\end{small}
\end{center}
\end{table*}

\textbf{IPM attack and partial participation.} To enhance the assessment of the proposed FedAA, we incorporate a more potent adversarial attack method known as Inner Product Manipulation (IPM) \citep{ipm}, which is crafted specifically to target Krum-based aggregation schemes. In \cref{ipmattack}, C=100\% signifies the participation of all clients in the aggregation phase, whereas C=50\% indicates that only a subset, specifically half, of the clients are involved in the process, characteristic of a partial participation FL framework.

Our findings demonstrate that the IPM presents a formidable challenge to the FedAvg method, though its impact is mitigated against more sophisticated defensive strategies. In every examined scenario, the FedAA consistently outpaces the benchmark models. Furthermore, the FedAA exhibited remarkable adaptability to scenarios with only partial client participation in the aggregation phase, and in certain cases, it even delivered superior performance.

\begin{table}[htbp]
\caption{Comparative analysis of three more challenging datasets.} 
\label{moresets}
\begin{center}
\begin{sc}
\begin{small}
\resizebox{\columnwidth}{!}{
\begin{tabular}{cccc} 
\hline
          & \multicolumn{3}{c}{datasets}                 \\ 
\cline{2-4}
Method    & CIFAR-100    & Tiny-iamgenet & AGnews-100    \\ 
\hline
FedAA     & \textbf{0.524(0.006)} & \textbf{0.260(0.021)}  & \textbf{0.955(0.126)}  \\
Ditto     & 0.239(0.007) & 0.203(0.029)  & 0.950(0.047)  \\
lp-proj-1 & 0.478(0.000)  & 0.224(0.000)  & 0.946(0.038)  \\
lp-proj-2 & 0.470(0.000) & 0.219(0.000)  & 0.946(0.039)  \\
fedavg    & 0.019(0.000) & 0.095(0.000)  & 0.312(0.142)  \\
\hline

\end{tabular}}
\end{small}
\end{sc}
\end{center}
\end{table}

\textbf{Challenging datasets.} To fully explore the capabilities of FedAA, we introduce three additional challenging datasets. The first, CIFAR-100 \citep{cifar10}, comprises 60,000 color images evenly distributed across 100 categories. The second, Tiny-ImageNet \citep{timgnet}, is a compact version of the ImageNet dataset, configured with 200 classes. Lastly, AG-News \citep{agnews}, offers a corpus exceeding one million news articles categorized into four distinctive sections. Each serves to benchmark text classification and machine learning models adept in natural language processing. As demonstrated in \cref{moresets}, FedAA surpasses Ditto's performance by approximately 30\%, 6\%, and 0.5\% on these datasets, respectively.

\begin{table}[htbp]
\caption{Comparative analysis of different size of held-out validation datasets.} 
\label{unfairdataset}
\begin{center}
\begin{small}
\begin{sc}
\resizebox{0.7\columnwidth}{!}{
\begin{tabular}{ccc} 
\hline
Datasets                      & Size       & Accuracy    \\
\hline
\multirow{4}{*}{Cifar10}      & 10     &  0.842(0.023)           \\
                              & 50     &  0.854(0.015)           \\
                              & 100    &  \textbf{0.875(0.024)}           \\
                              & unfair &  0.855(0.043)           \\ 
\hline
\multirow{4}{*}{Mnist}        & 10     &  0.976(0.001)           \\
                              & 50     &  0.976(0.001)           \\
                              & 100    &  \textbf{0.978(0.001)}           \\
                              & unfair &  0.978(0.001)           \\ 
\hline

\multirow{4}{*}{fashionmnist} & 10     &  0.972(0.025)           \\
                              & 50     &  0.974(0.022)           \\
                              & 100    &  \textbf{0.975(0.038)}           \\
                              & unfair &  0.975(0.047)           \\
\hline 
\end{tabular}}
\end{sc}
\end{small}
\end{center}
\end{table}

In \cref{unfairdataset}, we conduct comparative experiments to more effectively assess the performance of the held-out validation set. Starting with an example from Cifar-10, we modify the size of the set from 10 to 100 images for each category. In a specific scenario involving an unfair dataset, which includes 100 images each for categories 'airplane' and 'automobile' (labels 0 and 1, respectively), and only 10 images for all other categories. It is noted that an increase in the size of the dataset led to a modest improvement in accuracy. Crucially, the variance observed in this unfair dataset scenario is markedly higher compared to that in the other three scenarios.

\section{Conclusion and Discussion}\label{Conclusion}
In this paper, we model each communication round in the FL as an MDP and propose a simple framework FedAA, that seamlessly integrates DDPG into distributed learning with the capability of continuous action control. In addition, we reveal the tension between robustness and fairness at the server level. FedAA can simultaneously deliver superior performance, robustness, and comparable fairness. To attain this objective, we propose a novel client selection algorithm and offer the tradeoff by regulating the number $\rm M$ of clients participating in the aggregation. 

In future work, the integration of DRL into frameworks similar to Ditto and lp-proj could yield more robust models. Furthermore, through careful design of states, actions, and rewards, DRL can be applied to problems that are challenging for traditional approaches to handle. In addition, DRL may require a substantial number of transitions for training to unleash its optimal performance. Therefore, future endeavors could explore the pre-training of a robust DRL model capable of accurately identifying malicious clients, assigning them lower aggregation weights. Simultaneously, by leveraging the distance relationships between models, adaptive weight allocation can be achieved to yield fairer results. Therefore, the trade-off between the resulting overhead and performance enhancement is a noteworthy consideration for further contemplation.

%\begin{comment}

\section{Limitations.}\label{limitaions}
Due to the objective of maximizing cumulative rewards in DRL, this may result in non-convex optimization problems, rendering the training process susceptible to influences from factors such as initialization, and hyperparameter selection, especially evident when dealing with synthetic datasets. Another limitation is that, under strong attacks such as sign flipping, the proportion of malicious clients cannot exceed 50\% for FedAA, whereas lp-proj \citep{lp_proj} can tolerate up to 80\%. This difference arises from the fact that FedAA and lp-proj are two distinct frameworks. In lp-proj, it is possible to reduce the joint optimization problem to pure local training, allowing for a higher setting of malicious client numbers. 
    
%\end{comment}
\section{Acknowledgments}
This research was supported by National Key Research and Development Program of China under Grant 2022ZD0115301.

\bibliography{main}
\onecolumn
\section{Appendix / supplemental material}\label{implementation details}

\subsection{Experimental Setup}\label{experimental setup}
We conduct experiments on six datasets, MNIST \citep{mnist}, CIFAR10 \citep{cifar10}, FASHIONMNIST \citep{fashionmnist}, EMNIST \citep{emnist}, and two synthetic datasets \citep{synthetic}. For the first four real data sets, we extract 100 instances from each class, creating a held-out dataset at the server, ensuring fairness across all clients. The remaining dataset is divided into non-IID datasets for each client utilizing Dirichlet sampling with a parameter of $\alpha=0.1$. Further division into train/test sets is then performed on each client. For generating Synthetic datasets, we follow \citet{synthetic, fedprox, lp_proj}. The dataset is denoted as SYNTHETIC($\alpha,\beta$), where $\alpha$ controls the discrepancy of each client's model, while $\beta$ controls the discrepancy of each client's data. In this context, we generate SYNTHETIC(0,0) and SYNTHETIC(1,1). Specifically, for client $k$ with sample size $n_k$, the corresponding data samples $(X_k,Y_k)$ can be generated according to $y\ =\ argmax (\textrm{softmax}(W_kx+b_k))$ with $x\in \mathbb{R}^{60}$, $W_k\in\mathbb{R}^{10\times60}$ and $b_k\in\mathbb{R}^{10}$. We model $W_k\in \mathcal{N}(u_k,1)$, $b_k\in\mathcal{N}(u_k,1)$, $u_k,1\in\mathcal{N}(0,\alpha)$, and $(x_k)_j\in\mathcal{N}(v_k, \frac{1}{j^{1.2}})$ with $v_k\in \mathcal{N}(\mu_k,1)$ and $\mu_k\in\mathcal{N}(0,\beta)$. We set the parameter $\rm M$ to 30\% in experiments unless otherwise specified. 

\textbf{Model architectures}
The performance of FedAA and a suit of baselines are evaluated on both convex and non-convex models. In consideration of the scale of the datasets and the complexity of the tasks, we employed distinct neural network architectures to handle different datasets. Specifically, a 1-layer neural network and a 2-layer neural network are utilized for the MNIST and EMNIST datasets, respectively. In contrast, for the CIFAR10 and FASHIONMNIST datasets, we use convolutional neural networks. As for the SYNTHETIC datasets, a logistic regression model is employed. 

We give the details of the models as follows:
\begin{itemize}
    \item \textbf{1-hidden-layer neural network for MNIST.} The model consists of one fully connected (FC) layer with 100 neurons serving as the hidden layer, employing the ReLU as the activation function. In total, there are 79,510 parameters.
    \item \textbf{2-hidden-layer neural network for EMNIST.} The model consists of two fully connected layers with 100 neurons serving as the hidden layer. For each FC layer, we use the ReLU as the activation function. In total, there are 94,862 parameters.
    \item \textbf{CNN for CIFAR10.} The model consists of two convolutional layers and three fully connected layers. In total, there are 62,006 parameters. 
    The details are as follows:
    \begin{itemize}
    \item Convolutional layer 1: input channel: 3, output channel: 6, kernel size: 5.
    \item ReLU activation function, max pooling: kernel size: 2, stride: 2.
    \item Convolutional layer 2: input channel: 6, output channel: 16, kernel size: 5.
    \item ReLU activation function, max pooling: kernel size: 2, stride: 2.
    \item Fully connected layer 1: input features: 400, output features: 120.
    \item Fully connected layer 1: input features: 120, output features: 84.
    \item Fully connected layer 1: input features: 84, output features: 10.
    \end{itemize}
    \item \textbf{CNN for FASHIONMNIST.} The model we implement in we is modified from Resnet \citep{resnet}, which consists of one convolutional layer, two Resnet blocks, and a fully connected layer. In total, there are 678,794 parameters. 
    The details are as follows:
    \begin{itemize}
    \item Convolutional layer: input channel: 1, output channel: 64, kernel size: 7, stride: 2, padding: 3.
    \item Batch normalization, ReLU activation function, max pooling: kernel size: 3, stride: 2, padding: 1.
    \item Resnet block 1: input channel: 64, output channel: 64, 2 residuals.
    \item Resnet block 2: input channel: 64, output channel: 128, 2 residuals.
    \item Average pooling.
    \item Fully connected layer: input features: 128, output features: 10.    
    \end{itemize}
\end{itemize}
{In this study, the learning rate (lr) for the client is set to 0.1 with no weight decay. As for the actor and critic networks, their lr is set to 1e-2 with a weight decay of 1e-5. Additionally, the parameters for soft updating the target actor and critic networks are $\gamma$=0.99 and $\epsilon$=0.001. The batch size is fixed at 64, and the local model updates for 20 epochs.}

The details of the baselines are as follows:
\begin{itemize}
    \item \textbf{FedAvg} stands for the straightforward yet powerful approach within the FL. It enables clients to participate in a distributed learning process without direct data sharing. However, the robustness of FedAvg is relatively poor \citep{fedavg}.
    \item \textbf{Ditto} is the first framework that simultaneously offers robustness and fairness and mitigates the tension between the two criteria under a statistical heterogeneity scenario \citep{ditto}.
    \item \textbf{lp-proj} unifies the previous work on Ditto and projects the update process into lower dimensions. Depending on the different norms of the proximal term to the local subproblem, it can be categorized into lp-proj-1 and lp-proj-2 \citep{lp_proj}.
\end{itemize}

The details of the datasets are as follows:
\begin{table*}[h]
\caption{Summary of datasets and configurations. (The symbol $\triangle $ represents the data partitioning scheme within the synthetic dataset.  Specifically, we calculate the sample size for each client and denote the minimum as $n$. Then, each client uploads $0.1n$ samples to the server.)}
\label{datasets_models}

\begin{center}
\begin{small}
\begin{sc}
\resizebox{\textwidth}{!}{
\begin{tabular}{cccccc} 
\hline
dataset        & number of clients & M\%  & sample size at the server           & tasks                   & models              \\ 
\hline
mnist          & 100               & 30 & 10 categories, each with 100 images & 10-class classification & 1-hidden-layer NN   \\
cifar10        & 100               & 30 & 10 categories, each with 100 images & 10-class classification & cnn                 \\
fashionmnist   & 100               & 30 & 10 categories, each with 100 images & 10-class classification & cnn                 \\
emnist         & 300               & 30 & 62 categories, each with 100 images & 62-class classification & 2-hidden-layers NN  \\
synthetic(0,0) & 100               & 30 & $\triangle $                                & 10-class classification & logistic            \\
synthetic(1,1) & 100               & 30 & $\triangle $                                & 10-class classification & logistic            \\
\hline
\end{tabular}}
\end{sc}
\end{small}
\end{center}
\end{table*}

\textbf{Computing Resource.} We perform all our experiments on GPUs. One single experiment is performed on one single GPU. The details of the GPU are as follows:

\noindent $\bullet$ NVIDIA L40S with 48G memory, driver version: 535.129.03, CUDA version: 12.2.

\subsection{Full Results}\label{Full Results}
\begin{table}[h]
\caption{Mean test accuracy and variance for different percentages of aggregating clients participating in aggregation, under the condition of same value attack, on the CIFAR10 dataset. (The numbers within parentheses represent the variance.)}

\label{tradeoff_table}
\begin{center}
\begin{small}
\begin{sc}
\begin{tabular}{ccccc} 
\hline
      & \multicolumn{4}{c}{same value}                                                                 \\ 
\cline{2-5}
M(\%) & 10\%                  & 20\%                  & 30\%                  & 40\%                   \\ 
\hline
10    & 0.871(0.043)          & 0.879(0.039)          & 0.874(0.029)          & 0.870(0.041)           \\
20    & 0.879(0.029)          & 0.884(0.029)          & 0.880(0.030)          & 0.878(0.035)           \\
30    & 0.880(0.031)          & 0.889(0.031)          & 0.875(0.024)          & 0.876(0.030)           \\
40    & 0.889(0.026)          & 0.892(0.018)          & 0.889(0.025)          & 0.890(0.024)           \\
50    & 0.891(0.022)          & 0.894(0.017)          & 0.897(0.022)          & 0.895(0.021)           \\
60    & 0.892(0.021)          & 0.895(0.019)          & 0.895(0.021)          & \textbf{0.900(0.017)}  \\
70    & 0.895(0.022)          & 0.899(0.014)          & \textbf{0.902(0.016)} & 0.834(0.063)           \\
80    & 0.898(0.020)          & \textbf{0.903(0.013)} & 0.832(0.066)          & 0.838(0.064)           \\
90    & \textbf{0.900(0.015)} & 0.856(0.059)          & 0.838(0.063)          & 0.840(0.064)           \\
100   & 0.832(0.064)          & 0.857(0.058)          & 0.827(0.060)          & 0.836(0.064)           \\
\hline
\end{tabular}
\end{sc}
\end{small}
\end{center}

\end{table}

\begin{figure}[!h]

\begin{center}
\centerline{\includegraphics[width=\columnwidth]{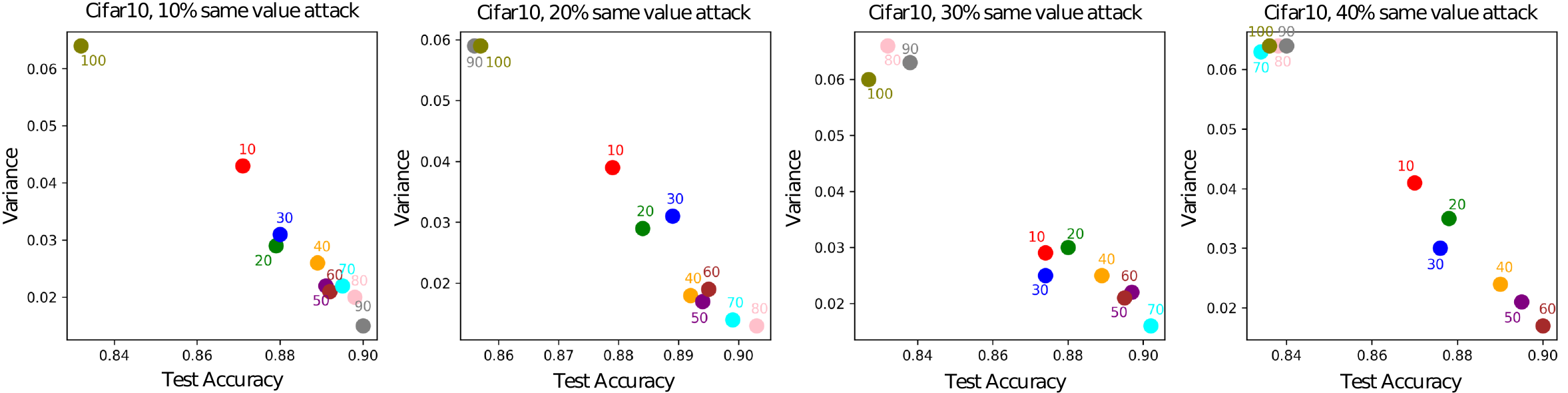}}
\caption{The tradeoff between test accuracy and fairness under the same value attack. The closer the $\rm M$ is to the lower right corner, the better. (The numbers in the figure correspond to the percentages participating in aggregation, e.g. 80 means $\rm M=80\%$)}
\label{tradeoff on M}
\end{center}

\end{figure}

\begin{table*}[h]
\caption{The remaining results of compression methods on different datasets.}
\begin{center}
\begin{small}
\begin{sc}
\resizebox{\textwidth}{!}{
\begin{tabular}{ccccccccccccc} 
\hline
    &              & mnist        &              &              & fashion-mnist &              &              & cifar10      &              &              & emnist       &               \\ 
\cline{2-13}
    & 256          & 128          & 64           & 256          & 128           & 64           & 256          & 128          & 64           & 256          & 128          & 64            \\ 
\hline
AL  & 0.978(0.001) & 0.979(0.001) & 0.978(0.000) & 0.975(0.038) & 0.966(0.068)  & 0.969(0.058) & 0.875(0.024) & 0.879(0.028) & 0.880(0.030) & 0.936(0.002) & 0.935(0.000) & 0.936(0.000)  \\
LHL & 0.978(0.003) & 0.980(0.002) & 0.981(0.002) & 0.974(0.038) & 0.969(0.038)  & 0.968(0.054) & 0.879(0.043) & 0.879(0.040) & 0.879(0.044) & 0.936(0.000) & 0.936(0.000) & 0.936(0.000)  \\
\hline
\end{tabular}}
\end{sc}
\end{small}
\end{center}
\end{table*}

\newpage

\begin{table}[htbp]
\caption{Complete test accuracy results under the attack of same-value. (The numbers within parentheses represent the variance.)}
\label{same-value}

\begin{center}
\begin{small}
\begin{sc}
\resizebox{\textwidth}{!}{
\begin{tabular}{cclccccc} 
\hline
Dataset                         & Methods   &  & clear                 & 10\%                  & 20\%                  & 30\%                  & 40\%                   \\ 
\cline{1-2}\cline{4-8}
\multirow{7}{*}{mnist}          & FedAvg    &  & 0.904(0.014)          & 0.124(0.054)          & 0.125(0.053)          & 0.132(0.053)          & 0.145(0.049)           \\
                                & local     &  & 0.965(0.007)          & 0.966(0.007)          & 0.965(0.008)          & 0.962(0.009)          & 0.962(0.010)           \\
                                & Global    &  & 0.874(0.101)          & 0.871(0.098)          & 0.859(0.093)          & 0.766(0.157)          & 0.784(0.126)           \\
                                & Ditto     &  & 0.972(0.005)          & 0.811(0.030)          & 0.778(0.033)          & 0.796(0.024)          & 0.769(0.027)           \\
                                & lp-proj-2 &  & 0.969(0.007)          & 0.967(0.006)          & 0.964(0.007)          & 0.968(0.006)          & 0.967(0.007)           \\
                                & lp-proj-1 &  & 0.971(0.007)          & 0.968(0.005)          & 0.969(0.005)          & 0.967(0.006)          & 0.969(0.007)           \\
                                & FedAA       &  & \textbf{0.977(0.006)} & \textbf{0.977(0.003)} & \textbf{0.979(0.002)} & \textbf{0.978(0.001)} & \textbf{0.971(0.006)}  \\ 
\hline
\multirow{7}{*}{CIFAR10}        & FedAvg    &  & 0.377(0.048)          & 0.113(0.062)          & 0.120(0.065)          & 0.133(0.067)          & 0.149(0.070)           \\
                                & local     &  & 0.852(0.028)          & 0.852(0.028)          & 0.871(0.017)          & 0.866(0.018)          & 0.866(0.019)           \\
                                & Global    &  & 0.276(0.152)          & 0.299(0.136)          & 0.319(0.157)          & 0.297(0.172)          & 0.351(0.101)           \\
                                & Ditto     &  & 0.867(0.019) & 0.817(0.048)          & 0.824(0.044)          & 0.815(0.043)          & 0.820(0.042)           \\
                                & lp-proj-2 &  & 0.866(0.021)          & 0.845(0.027)          & 0.854(0.016)          & 0.830(0.025)          & 0.852(0.021)           \\
                                & lp-proj-1 &  & 0.867(0.021) & 0.851(0.024)          & 0.834(0.026)          & 0.847(0.018)          & 0.859(0.014)           \\
                                & FedAA       &  & \textbf{0.873(0.015)} & \textbf{0.880(0.031)} & \textbf{0.889(0.031)}          & \textbf{0.874(0.025)} & \textbf{0.876(0.030)}  \\ 
\hline
\multirow{7}{*}{fashionmnist}   & FedAvg    &  & 0.762(0.069)          & 0.152(0.052)          & 0.167(0.053)          & 0.187(0.053)          & 0.190(0.052)           \\
                                & local     &  & 0.966(0.010)          & 0.963(0.010)          & 0.965(0.011)          & 0.966(0.011)          & 0.967(0.012)           \\
                                & Global    &  & 0.442(0.201)          & 0.424(0.208)          & 0.555(0.183)          & 0.562(0.215)          & 0.561(0.213)           \\
                                & Ditto     &  & \textbf{0.973(0.003)} & 0.814(0.045)          & 0.803(0.046)          & 0.829(0.048)          & 0.858(0.048)           \\
                                & lp-proj-2 &  & 0.969(0.004)          & 0.964(0.004)          & 0.964(0.003)          & 0.966(0.004)          & 0.964(0.005)           \\
                                & lp-proj-1 &  & 0.969(0.004)          & 0.966(0.004)          & 0.966(0.003)          & 0.968(0.004)          & 0.967(0.004)           \\
                                & FedAA       &  & 0.972(0.003)          & \textbf{0.968(0.026)} & \textbf{0.970(0.028)} & \textbf{0.975(0.038)} & \textbf{0.972(0.019)}  \\ 
\hline
\multirow{7}{*}{emnist}         & FedAvg    &  & 0.666(0.032)          & 0.054(0.014)          & 0.057(0.014)          & 0.062(0.014)          & 0.059(0.013)           \\
                                & local     &  & 0.853(0.018)          & 0.852(0.018)          & 0.853(0.018)          & 0.852(0.018)          & 0.853(0.018)           \\
                                & Global    &  & 0.710(0.010)          & 0.711(0.010)          & 0.732(0.010)          & 0.741(0.010)          & 0.762(0.010)           \\
                                & Ditto     &  & 0.881(0.003)          & 0.487(0.003)          & 0.478(0.003)          & 0.502(0.001)          & 0.509(0.001)           \\
                                & lp-proj-2 &  & 0.904(0.002)          & 0.863(0.002)          & 0.802(0.002)          & 0.884(0.001)          & 0.888(0.001)           \\
                                & lp-proj-1 &  & 0.901(0.002)          & 0.874(0.002)          & 0.875(0.002)          & 0.884(0.001)          & 0.888(0.001)           \\
                                & FedAA       &  & \textbf{0.934(0.003)} & \textbf{0.936(0.003)} & \textbf{0.936(0.002)} & \textbf{0.939(0.002)} & \textbf{0.941(0.002)}  \\ 
\hline
\multirow{7}{*}{synthetic(0,0)} & FedAvg    &  & 0.749(0.091)          & 0.553(0.139)          & 0.499(0.145)          & 0.431(0.129)          & 0.402(0.124)           \\
                                & local     &  & 0.877(0.017)          & 0.873(0.018)          & 0.871(0.018)          & 0.857(0.018)          & 0.857(0.019)           \\
                                & Global    &  & 0.470(0.064)          & 0.489(0.069)          & 0.523(0.074)          & 0.560(0.084))         & 0.570(0.086)           \\
                                & Ditto     &  & 0.851(0.018)          & 0.851(0.018)          & 0.859(0.015)          & 0.861(0.014)          & 0.869(0.014)           \\
                                & lp-proj-2 &  & \textbf{0.940(0.002)} & 0.916(0.004)          & 0.918(0.004)          & 0.911(0.005)          & 0.920(0.004)           \\
                                & lp-proj-1 &  & 0.924(0.004)          & 0.914(0.005)          & 0.919(0.004)          & 0.911(0.005)          & 0.918(0.004)           \\
                                & FedAA       &  & 0.932(0.163)          & \textbf{0.937(0.160)} & \textbf{0.939(0.148)} & \textbf{0.946(0.135)} & \textbf{0.943(0.159)}  \\ 
\hline
\multirow{7}{*}{synthetic(1,1)} & FedAvg    &  & 0.741(0.127)          & 0.533(0.168)          & 0.419(0.154)          & 0.381(0.140)          & 0.362(0.133)           \\
                                & local     &  & 0.899(0.025)          & 0.901(0.025)          & 0.906(0.024)          & 0.890(0.026)          & 0.894(0.029)           \\
                                & Global    &  & 0.160(0.046)          & 0.185(0.051)          & 0.195(0.056)          & 0.210(0.055)          & 0.244(0.063)           \\
                                & Ditto     &  & 0.879(0.021)          & 0.875(0.021)          & 0.871(0.021)          & 0.878(0.018)          & 0.869(0.019)           \\
                                & lp-proj-2 &  & \textbf{0.949(0.013)} & 0.910(0.016)          & 0.911(0.015)          & 0.909(0.017)          & 0.909(0.017)           \\
                                & lp-proj-1 &  & 0.945(0.013)          & 0.915(0.016)          & 0.925(0.015)          & 0.923(0.017)          & 0.920(0.018)           \\
                                & FedAA       &  & 0.936(0.147)          & \textbf{0.937(0.147)} & \textbf{0.938(0.140)} & \textbf{0.937(0.160)} & \textbf{0.920(0.140)}  \\
\hline
\end{tabular}}
\end{sc}
\end{small}
\end{center}

\end{table}

\newpage

\begin{table*}[!h]
\caption{Tuning the parameter learning rate of actor networks and critic networks within three RL algorithms.}
\begin{center}
\begin{small}
\begin{sc}
\resizebox{\textwidth}{!}{
\begin{tabular}{ccccccc} 
\hline
     &              & cifar10       &              &              & mnist         &               \\ 
\cline{2-7}
     &              & Learning rate &              &              & Learning rate &               \\
     & 1e-1        & 1e-2      & 1e-3     & 1e-1     & 1e-2      & 1e-3      \\ 
\hline
DDPG & 0.879(0.025) & 0.880(0.030)  & 0.884(0.023) & 0.965(0.001) & 0.979(0.001)  & 0.969(0.001)  \\
PPO  & 0.883(0.026) & 0.881(0.024)  & 0.881(0.021) & 0.962(0.001) & 0.964(0.001)  & 0.962(0.001)  \\
TD3  & 0.810(0.029) & 0.812(0.034)  & 0.811(0.030) & 0.939(0.000) & 0.941(0.000)  & 0.932(0.000)  \\
\hline
\end{tabular}}
\end{sc}
\end{small}
\end{center}
\end{table*}

\begin{table}[htbp]
\caption{Complete test accuracy results under the attack of sign-flipping. (The numbers within parentheses represent the variance, and $\star$ on the cell means in that case, the model will collapse.)}
\label{sign-flipping}

\begin{center}
\begin{small}
\begin{sc}
\resizebox{\textwidth}{!}{
\begin{tabular}{cclccccc} 
\hline
dataset                         & methods   &  & clear                 & 10\%                  & 20\%                  & 30\%                  & 40\%                   \\ 
\cline{1-2}\cline{4-8}
\multirow{7}{*}{mnist}          & FedAvg    &  & 0.904(0.014)          & 0.395(0.052)          & 0.511(0.048)          & $\star$              & $\star$               \\
                                & local     &  & 0.965(0.007)          & 0.966(0.007)          & 0.965(0.008)          & 0.962(0.009)          & 0.962(0.010)           \\
                                & Global    &  & 0.874(0.101)          & 0.729(0.203)          & 0.808(0.102)          & 0.863(0.078)          & 0.763(0.140)           \\
                                & Ditto     &  & 0.972(0.005)          & 0.930(0.010)          & 0.948(0.010)          & $\star$              & $\star$               \\
                                & lp-proj-2 &  & 0.969(0.008)          & 0.971(0.008)          & 0.972(0.008)          & $\star$              & $\star$               \\
                                & lp-proj-1 &  & 0.971(0.007)          & 0.969(0.008)          & 0.971(0.007)          & 0.968(0.007)          & 0.969(0.006)           \\
                                & FedAA       &  & \textbf{0.977(0.005)} & \textbf{0.972(0.010)} & \textbf{0.973(0.006)} & \textbf{0.974(0.002)} & \textbf{0.981(0.03)}   \\ 
\hline
\multirow{7}{*}{CIFAR10}        & FedAvg    &  & 0.377(0.048)          & 0.115(0.057)          & 0.123(0.055)          & $\star$              & $\star$               \\
                                & local     &  & 0.852(0.028)          & 0.852(0.028)          & \textbf{0.871(0.017)} & 0.866(0.018)          & 0.866(0.019)           \\
                                & Global    &  & 0.276(0.152)          & 0.293(0.157)          & 0.319(0.176)          & 0.320(0.152)          & 0.331(0.138)           \\
                                & Ditto     &  & {0.867(0.019)} & 0.860(0.022)          & 0.854(0.024)          & $\star$              & $\star$               \\
                                & lp-proj-2 &  & 0.866(0.021)          & 0.859(0.021)          & 0.857(0.019)          & $\star$              & $\star$               \\
                                & lp-proj-1 &  & {0.867(0.021)} & 0.858(0.022)          & 0.860(0.020)          & 0.850(0.023)          & 0.853(0.017)           \\
                                & FedAA       &  & \textbf{0.873(0.015)} & \textbf{0.872(0.031)} & \textbf{0.871(0.025)} & \textbf{0.875(0.025)} & \textbf{0.876(0.030)}  \\ 
\hline
\multirow{7}{*}{fashionmnist}   & FedAvg    &  & 0.762(0.069)          & 0.139(0.053)          & 0.172(0.052)          & $\star$              & $\star$               \\
                                & local     &  & 0.966(0.010)          & 0.963(0.010)          & 0.965(0.011)          & 0.966(0.011)          & 0.967(0.012)           \\
                                & Global    &  & 0.442(0.201)          & 0.449(0.208)          & 0.439(0.183)          & 0.578(0.230)          & 0.658(0.214)           \\
                                & Ditto     &  & \textbf{0.973(0.003)} & 0.956(0.006)          & 0.939(0.015)          & $\star$              & $\star$               \\
                                & lp-proj-2 &  & 0.969(0.004)          & 0.968(0.004)          & 0.968(0.003)          & $\star$              & $\star$               \\
                                & lp-proj-1 &  & 0.969(0.004)          & 0.967(0.004)          & 0.967(0.003)          & 0.966(0.004)          & 0.965(0.005)           \\
                                & FedAA       &  & 0.972(0.003)          & \textbf{0.969(0.020)} & \textbf{0.971(0.031)} & \textbf{0.971(0.023)} & \textbf{0.972(0.026)}  \\ 
\hline
\multirow{7}{*}{emnist}         & FedAvg    &  & 0.666(0.032)          & 0.055(0.012)          & 0.058(0.011)          & $\star$              & $\star$               \\
                                & local     &  & 0.853(0.018)          & 0.852(0.018)          & 0.853(0.018)          & 0.852(0.018)          & 0.853(0.018)           \\
                                & Global    &  & 0.710(0.010)          & 0.732(0.010)          & 0.743(0.010)          & 0.756(0.010)          & 0.767(0.010)           \\
                                & Ditto     &  & 0.881(0.003)          & 0.767(0.003)          & 0.718(0.003)          & $\star$              & $\star$               \\
                                & lp-proj-2 &  & 0.904(0.002)          & 0.902(0.002)          & 0.904(0.002)          & $\star$              & $\star$               \\
                                & lp-proj-1 &  & 0.901(0.002)          & 0.902(0.002)          & 0.905(0.002)          & 0.905(0.001)          & 0.903(0.001)           \\
                                & FedAA       &  & \textbf{0.934(0.003)} & \textbf{0.934(0.003)} & \textbf{0.938(0.002)} & \textbf{0.939(0.002)} & \textbf{0.940(0.002)}  \\ 
\hline
\multirow{7}{*}{synthetic(0,0)} & FedAvg    &  & 0.749(0.091)          & 0.362(0.079)          & 0.321(0.071)          & 0.128(0.057)          & $\star$               \\
                                & local     &  & 0.877(0.017)          & 0.873(0.018)          & 0.871(0.018)          & 0.857(0.018)          & 0.857(0.019)           \\
                                & Global    &  & 0.470(0.064)          & 0.489(0.069)          & 0.523(0.074)          & 0.560(0.084))         & 0.565(0.092)           \\
                                & Ditto     &  & 0.851(0.018)          & 0.849(0.018)          & 0.859(0.016)          & 0.832(0.023)          & $\star$               \\
                                & lp-proj-2 &  & \textbf{0.940(0.002)} & 0.922(0.004)          & 0.925(0.003)          & 0.869(0.013)          & $\star$               \\
                                & lp-proj-1 &  & 0.924(0.004)          & 0.922(0.004)          & \textbf{0.940(0.003)} & 0.857(0.015)          & 0.884(0.009)           \\
                                & FedAA       &  & 0.932(0.163)          & \textbf{0.937(0.160)} & 0.939(0.146)          & \textbf{0.945(0.134)} & \textbf{0.950(0.131)}  \\ 
\hline
\multirow{7}{*}{synthetic(1,1)} & FedAvg    &  & 0.741(0.127)          & 0.317(0.089)          & 0.236(0.079)          & 0.113(0.046)          & $\star$               \\
                                & local     &  & 0.899(0.025)          & 0.901(0.025)          & 0.906(0.024)          & 0.890(0.026)          & 0.894(0.029)           \\
                                & Global    &  & 0.160(0.046)          & 0.185(0.051)          & 0.195(0.056)          & 0.210(0.055)          & 0.244(0.063)           \\
                                & Ditto     &  & 0.879(0.021)          & 0.878(0.022)          & 0.871(0.022)          & 0.854(0.030)          & $\star$               \\
                                & lp-proj-2 &  & \textbf{0.949(0.013)} & 0.920(0.008)          & 0.911(0.009)          & 0.862(0.024)          & $\star$               \\
                                & lp-proj-1 &  & 0.945(0.013)          & 0.914(0.010)          & 0.910(0.009)          & 0.866(0.019)          & 0.876(0.013)           \\
                                & FedAA       &  & 0.936(0.147)          & \textbf{0.938(0.133)} & \textbf{0.935(0.153)} & \textbf{0.938(0.161)} & \textbf{0.939(0.169)}  \\
\hline
\end{tabular}}
\end{sc}
\end{small}
\end{center}

\end{table}

\newpage

\begin{table*}[!h]
\caption{Tuning the parameter $\gamma$ within three RL algorithms.}
\begin{center}
\begin{small}
\begin{sc}
\resizebox{\textwidth}{!}{
\begin{tabular}{ccccccc} 
\hline
     &              & cifar10      &              &              & mnist        &               \\ 
\cline{2-7}
     &              &$ \gamma $        &              &              & $\gamma $        &               \\
     & 0.9          & 0.99         & 0.999        & 0.9          & 0.99         & 0.999         \\ 
\hline
DDPG & 0.878(0.026) & 0.880(0.030) & 0.874(0.025) & 0.970(0.001) & 0.979(0.001) & 0.971(0.001)  \\
PPO  & 0.880(0.021) & 0.881(0.024) & 0.879(0.034) & 0.963(0.001) & 0.964(0.001) & 0.963(0.001)  \\
TD3  & 0.809(0.034) & 0.812(0.034) & 0.812(0.034) & 0.938(0.001) & 0.941(0.000) & 0.942(0.000)  \\
\hline
\end{tabular}}
\end{sc}
\end{small}
\end{center}
\end{table*}

\begin{table}[htbp]
\caption{Complete test accuracy results under the attack of Gaussian. (The numbers within parentheses represent the variance.)}
\label{gaussian}

\begin{center}
\begin{small}
\begin{sc}
\resizebox{\textwidth}{!}{
\begin{tabular}{cclccccc} 
\hline
dataset                         & methods   &  & clear                 & 10\%                  & 20\%                  & 30\%                  & 40\%                   \\ 
\cline{1-2}\cline{4-8}
\multirow{7}{*}{mnist}          & FedAvg    &  & 0.904(0.014)          & 0.216(0.020)          & 0.208(0.021)          & 0.203(0.019)          & 0.205(0.022)           \\
                                & local     &  & 0.965(0.007)          & 0.966(0.007)          & 0.965(0.008)          & 0.962(0.009)          & 0.962(0.010)           \\
                                & Global    &  & 0.874(0.101)          & 0.744(0.134)          & 0.773(0.132)          & 0.786(0.120)          & 0.824(0.122)           \\
                                & Ditto     &  & 0.972(0.005)          & 0.881(0.013)          & 0.879(0.015)          & 0.869(0.013)          & 0.862(0.014)           \\
                                & lp-proj-2 &  & 0.969(0.008)          & 0.960(0.008)          & 0.961(0.008)          & 0.959(0.008)          & 0.953(0.009)           \\
                                & lp-proj-1 &  & 0.971(0.007)          & 0.962(0.007)          & 0.962(0.010)          & 0.962(0.008)          & 0.963(0.008)           \\
                                & FedAA       &  & \textbf{0.977(0.005)} & \textbf{0.973(0.008)} & \textbf{0.972(0.009)} & \textbf{0.973(0.006)} & \textbf{0.977(0.003)}  \\ 
\hline
\multirow{7}{*}{CIFAR10}        & FedAvg    &  & 0.377(0.048)          & 0.131(0.018)          & 0.146(0.024)          & 0.147(0.026)          & 0.158(0.029)           \\
                                & local     &  & 0.852(0.028)          & 0.852(0.028)          & \textbf{0.871(0.017)} & 0.866(0.018)          & 0.866(0.019)           \\
                                & Global    &  & 0.276(0.152)          & 0.290(0.181)          & 0.309(0.157)          & 0.331(0.146)          & 0.311(0.146)           \\
                                & Ditto     &  & {0.867(0.019)} & 0.814(0.048)          & 0.815(0.045)          & 0.815(0.042)          & 0.820(0.042)           \\
                                & lp-proj-2 &  & 0.866(0.021)          & 0.822(0.034)          & 0.822(0.033)          & 0.821(0.024)          & 0.820(0.025)           \\
                                & lp-proj-1 &  & {0.867(0.021)} & 0.823(0.031)          & 0.822(0.032)          & 0.819(0.026)          & 0.820(0.026)           \\
                                & FedAA       &  & \textbf{0.873(0.015)} & \textbf{0.872(0.030)} & \textbf{0.871(0.026)} & \textbf{0.876(0.028)} & \textbf{0.876(0.024)}  \\ 
\hline
\multirow{7}{*}{fashionmnist}   & FedAvg    &  & 0.762(0.069)          & 0.183(0.046)          & 0.205(0.044)          & 0.181(0.047)          & 0.174(0.044)           \\
                                & local     &  & 0.966(0.010)          & 0.963(0.010)          & 0.965(0.011)          & 0.966(0.011)          & 0.967(0.012)           \\
                                & Global    &  & 0.442(0.201)          & 0.395(0.208)          & 0.408(0.218)          & 0.545(0.172)          & 0.634(0.210)           \\
                                & Ditto     &  & \textbf{0.973(0.003)} & 0.814(0.044)          & 0.803(0.056)          & 0.827(0.049)          & 0.853(0.057)           \\
                                & lp-proj-2 &  & 0.969(0.004)          & 0.960(0.005)          & 0.968(0.004)          & 0.965(0.004)          & 0.960(0.005)           \\
                                & lp-proj-1 &  & 0.969(0.004)          & 0.959(0.005)          & 0.959(0.004)          & 0.964(0.004)          & 0.960(0.005)           \\
                                & FedAA       &  & 0.972(0.003)          & \textbf{0.967(0.024)} & \textbf{0.970(0.020)} & \textbf{0.971(0.021)} & \textbf{0.972(0.024)}  \\ 
\hline
\multirow{7}{*}{emnist}         & FedAvg    &  & 0.666(0.032)          & 0.063(0.001)          & 0.064(0.001)          & 0.058(0.002)          & 0.057(0.001)           \\
                                & local     &  & 0.853(0.018)          & 0.852(0.018)          & 0.853(0.018)          & 0.852(0.018)          & 0.853(0.018)           \\
                                & Global    &  & 0.710(0.010)          & 0.730(0.010)          & 0.726(0.010)          & 0.744(0.008)          & 0.771(0.010)           \\
                                & Ditto     &  & 0.881(0.003)          & 0.618(0.003)          & 0.599(0.003)          & 0.584(0.001)          & 0.582(0.001)           \\
                                & lp-proj-2 &  & 0.904(0.002)          & 0.588(0.002)          & 0.588(0.002)          & 0.561(0.001)          & 0.581(0.001)           \\
                                & lp-proj-1 &  & 0.901(0.002)          & 0.851(0.002)          & 0.874(0.002)          & 0.881(0.001)          & 0.883(0.001)           \\
                                & FedAA       &  & \textbf{0.934(0.003)} & \textbf{0.936(0.003)} & \textbf{0.938(0.002)} & \textbf{0.937(0.002)} & \textbf{0.941(0.002)}  \\ 
\hline
\multirow{7}{*}{synthetic(0,0)} & FedAvg    &  & 0.749(0.091)          & 0.191(0.045)          & 0.176(0.045)          & 0.184(0.042)          & 0.189(0.041)           \\
                                & local     &  & 0.877(0.017)          & 0.873(0.018)          & 0.871(0.018)          & 0.857(0.018)          & 0.857(0.019)           \\
                                & Global    &  & 0.470(0.064)          & 0.489(0.069)          & 0.523(0.074)          & 0.560(0.084))         & 0.565(0.092)           \\
                                & Ditto     &  & 0.851(0.018)          & 0.795(0.039)          & 0.803(0.035)          & 0.814(0.032)          & 0.814(0.036)           \\
                                & lp-proj-2 &  & \textbf{0.940(0.002)} & 0.897(0.007)          & 0.898(0.007)          & 0.902(0.007)          & 0.906(0.005)           \\
                                & lp-proj-1 &  & 0.924(0.004)          & 0.892(0.007)          & 0.896(0.007)          & 0.893(0.008)          & 0.902(0.007)           \\
                                & FedAA       &  & 0.932(0.163)          & \textbf{0.936(0.159)} & \textbf{0.939(0.145)} & \textbf{0.952(0.165)} & \textbf{0.950(0.132)}  \\ 
\hline
\multirow{7}{*}{synthetic(1,1)} & FedAvg    &  & 0.741(0.127)          & 0.181(0.051)          & 0.189(0.059)          & 0.195(0.054)          & 0.206(0.056)           \\
                                & local     &  & 0.899(0.025)          & 0.901(0.025)          & 0.906(0.024)          & 0.890(0.026)          & 0.894(0.029)           \\
                                & Global    &  & 0.160(0.046)          & 0.185(0.051)          & 0.195(0.056)          & 0.210(0.055)          & 0.244(0.063)           \\
                                & Ditto     &  & 0.879(0.021)          & 0.847(0.040)          & 0.834(0.042)          & 0.820(0.044)          & 0.808(0.045)           \\
                                & lp-proj-2 &  & \textbf{0.949(0.013)} & 0.898(0.012)          & 0.896(0.011)          & 0.898(0.011)          & 0.896(0.011)           \\
                                & lp-proj-1 &  & 0.945(0.013)          & 0.897(0.013)          & 0.903(0.010)          & 0.891(0.012)          & 0.897(0.011)           \\
                                & FedAA       &  & 0.936(0.147)          & \textbf{0.934(0.148)} & \textbf{0.935(0.150)} & \textbf{0.942(0.138)} & \textbf{0.939(0.157)}  \\
\hline

\end{tabular}}
\end{sc}
\end{small}
\end{center}

\end{table}

%\end{comment}
\end{document}